\documentclass[10pt,twocolumn,letterpaper]{article}

\usepackage[pagenumbers]{cvpr} 

\usepackage[dvipsnames]{xcolor}

\usepackage{cuted}

\usepackage{bm}
\usepackage{amsmath}
\usepackage{amssymb}

\definecolor{cvprblue}{rgb}{0.21,0.49,0.74}

\PassOptionsToPackage{pagebackref,breaklinks,colorlinks,citecolor=cvprblue}{hyperref}
\usepackage[a-1b]{pdfx}

\title{Arc2Avatar: Generating Expressive 3D Avatars from a Single Image \\via ID Guidance}

\author{
Dimitrios Gerogiannis\quad
Foivos Paraperas Papantoniou\quad
Rolandos Alexandros Potamias\\
Alexandros Lattas\quad
Stefanos Zafeiriou\\[6pt]
Imperial College London, UK\\
{\tt\small \{d.gerogiannis22, f.paraperas, r.potamias, a.lattas, s.zafeiriou\}@imperial.ac.uk}
}

\begin{document}
\maketitle

\begin{strip}
\centering
\includegraphics[width=\textwidth]{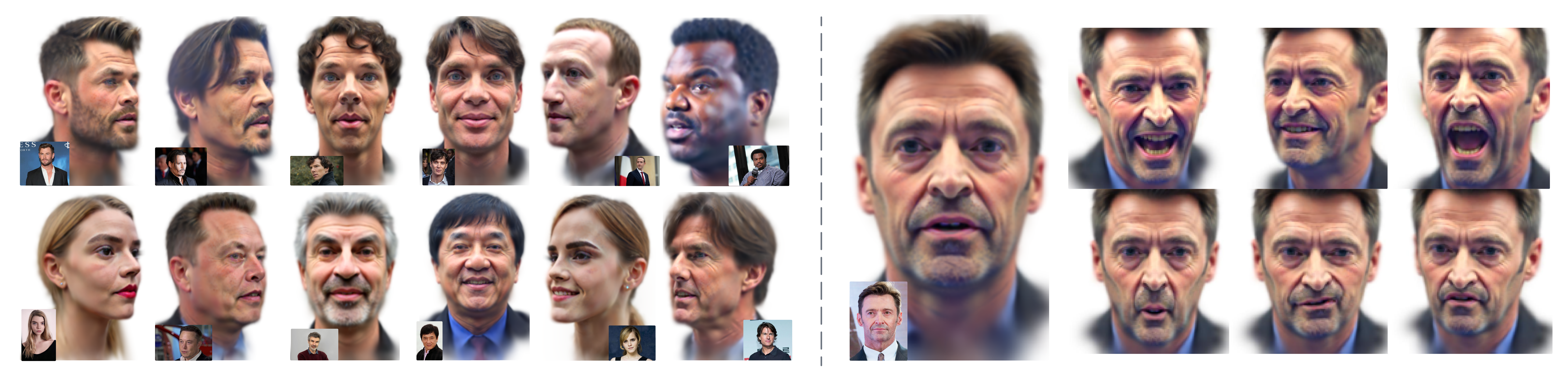}
\captionof{figure}{Arc2Avatar creates detailed 3D head avatars from a single image with unprecedented realism and identity similarity through a carefully designed score distillation sampling approach on top of an adapted 2D face foundation model. Our method supports blendshape-driven expression generation by retaining dense correspondence between the 3D Gaussian Splats and an underlying facial mesh template.}
\label{fig:teaser}
\end{strip}

\begin{abstract}
 Inspired by the effectiveness of 3D Gaussian Splatting (3DGS) in reconstructing detailed 3D scenes within multi-view setups and the emergence of large 2D human foundation models, we introduce Arc2Avatar, the first SDS-based method utilizing a human face foundation model as guidance with just a single image as input. To achieve that, we extend such a model for diverse-view human head generation by fine-tuning on synthetic data and modifying its conditioning. Our avatars maintain a dense correspondence with a human face mesh template, allowing blendshape-based expression generation. This is achieved through a modified 3DGS approach, connectivity regularizers, and a strategic initialization tailored for our task. Additionally, we propose an optional efficient SDS-based correction step to refine the blendshape expressions, enhancing realism and diversity. Experiments demonstrate that Arc2Avatar achieves state-of-the-art realism and identity preservation, effectively addressing color issues by allowing the use of very low guidance, enabled by our strong identity prior and initialization strategy, without compromising detail. Please visit our \href{https://arc2avatar.github.io/}{\textcolor{magenta}{project page}} for more resources.
\end{abstract}

\section{Introduction}
\label{sec:intro}
Recent advances in 3D representations have significantly transformed the field of 3D avatar generation. For many years, 3D Morphable Models (3DMMs) \cite{blanzmorphable,blanz2003face,FLAME:SiggraphAsia2017,ploumpis2019combining,ploumpis2020towards} have been the standard solution, offering precise control over avatars through manipulation of the underlying meshes and a decent degree of identity representation through PCA. However, 3DMMs often struggle to capture fine details, such as the complexity of hair, limiting their ability to represent the entirety of the head. The emergence of Neural Radiance Fields (NeRFs) \cite{mildenhall2020nerfrepresentingscenesneural}, opened new possibilities by implicitly modeling complex geometries and textures present in human heads \cite{Kirschstein_2023, I_k_2023, gafni2020dynamicneuralradiancefields, yu2023nofanerfbasedoneshotfacial} via MLPs, however their representation is computationally intensive and difficult to manipulate.
Recently, 3D Gaussian Splatting (3DGS) \cite{kerbl3Dgaussians} has emerged as a groundbreaking solution for 3D scene representation. Characterized by real-time performance and enhanced detail capturing abilities, 3DGS has been applied to represent 3D head avatars. Many methods aim to leverage the controllability of 3DMMs by embedding them into the splatting process \cite{qian2024gaussianavatarsphotorealisticheadavatars, xu2024gaussianheadavatarultra, shao2024splattingavatarrealisticrealtimehuman, xiang2024flashavatarhighfidelityheadavatar}, allowing flexible control over the representation. Although these approaches have shown impressive results, they rely on multi-view input data of a subject, which limits their applicability compared to single-image reconstruction approaches.

Parallel to these developments, the field of generative modeling has witnessed significant advancements, especially with Latent Diffusion Models \cite{ho2020denoising, rombach2022high}. Large foundation text-to-image models have been proposed, that can additionally serve as robust priors for various tasks. However, their utilization in the 3D space is challenging. The seminal work of \cite{poole2022dreamfusion} bridged this gap by introducing Score Distillation Sampling (SDS), allowing 3D assets to be optimized to match textual descriptions. Since then, SDS has become a potent approach in conditional 3D asset generation, primarily from text, and has been applied to avatar generation as well \cite{huang2023humannorm, han2023headsculptcrafting3dhead, liu2024headartisttextconditioned3dhead, chen2023fantasia3d, zhou2024headstudiotextanimatablehead, liao2023tada, cao2023dreamavatartextandshapeguided3d}. Some of these methods even extend to animatable avatars but often suffer from low fidelity \cite{liao2023tada, zhou2024headstudiotextanimatablehead}.

A major issue shared among SDS-based approaches is oversaturation and unnatural colors, stemming from unstable SDS gradients. This instability results in avatars that appear overly saturated and unrealistic. 
Additionally, relying on text as guidance for avatar generation presents its own challenges, as textual descriptions are often vague and lack sufficient identity-specific information. This can lead to distortions and artifacts beyond color issues, affecting the fidelity and accuracy of the generated avatars. 

Motivated by the scarcity of methods in the SDS literature that generate highly realistic, expressive 3D avatars from a single image, we present our approach, termed Arc2Avatar, which addresses these challenges directly by leveraging robust priors from human face foundation models, the representational power of 3DGS, and the controllability of 3DMMs. We use Arc2Face \cite{paraperas2024arc2face}, a state-of-the-art human face foundation model conditioned on identity embeddings rather than text embeddings. However, to overcome Arc2Face's predominantly frontal focus, we extend it for diverse-view generation using synthetic data and Low Rank Adaptation \cite{hu2022lora} (LoRA) for efficient fine-tuning. Our method employs a modified SDS approach, where splats are strategically initialized on a densely sampled FLAME 3DMM \cite{FLAME:SiggraphAsia2017} with a masked 3DGS setup. In this framework, splats in the facial area are protected from densification, pruning, and opacity resets, ensuring they maintain consistent alignment with the template. This alignment is further reinforced by template-based regularizers, which preserve proximity to the template mesh throughout optimization.

The proposed approach overcomes the limitation of SDS approaches to produce detailed 3D assets with natural colors due to the inevitable uncertainty in text prompts, by leveraging the precise ID-guidance of Arc2Face, along with template-based regularizers, targeted initialization, and a masked strategy for facial splats. Additionally, this allows for the use of reduced diffusion guidance scale during distillation, in contrast to previous Stable Diffusion-based \cite{rombach2022high} distillation methods.
Arc2Avatar generates complete 3D heads with unprecedented realism, detail, and natural color fidelity. By leveraging the dense template correspondence, our method further enables realistic expressions through 3DMM blendshapes (see \cref{fig:teaser}). Finally, we introduce an optional SDS-based refinement step to correct exaggerated expressions and enhance expression fidelity. In summary, the contributions of this paper are:
\begin{itemize}
    \item The first SDS-based method to successfully generate realistic 3D head avatars from a single image using a face foundation model as guidance.
    \item A carefully designed, task-specific SDS process effective for ID-driven 3D avatar generation.
    \item Blendshape-enabled 3DGS avatars powered by a 3DMM that allows expression generation and refinement using the same framework.
\end{itemize}

\section{Related Work}
\label{sec:related}

\subsection{ID-Conditioned Diffusion Models}
\label{sec:id_conditioned_generation}
Following the remarkable success of text-to-image diffusion models, recent research has focused on adapting foundation models, such as Stable Diffusion (SD)~\cite{rombach2022high}, for subject-specific image generation. Early methods, including Textual Inversion \cite{gal2022image} and DreamBooth \cite{ruiz2023dreambooth}, leveraged model or embedding optimization to customize SD based on a few images of a target object. However, these methods often encounter challenges due to lengthy optimization times. Subsequent works ~\cite{gal2023encoder, zhou2023enhancing} introduced encoders for direct subject conditioning, while ~\cite{kumari2023multi} reduced optimization demands by fine-tuning only a subset of the model’s parameters. Although these methods can efficiently preserve the characteristics of arbitrary objects, their ability to adapt to human subjects is limited by the lack of identity-specific features in their conditioning inputs.

To achieve consistent identity preservation in human image generation, more recent approaches incorporate facial features to condition text-to-image models. Within this framework, the CLIP image encoder \cite{radford2021learning} has become a popular choice for extracting features from facial images, as demonstrated in FastComposer \cite{xiao2023fastcomposer}, PhotoVerse \cite{chen2023photoverse}, MoA \cite{wang2024moa}, and PhotoMaker \cite{li2023photomaker}. These features are then introduced to the pre-trained SD model as a supplement to text embeddings through cross-attention layers. Alternatively, Celeb-Basis \cite{yuan2023inserting} and StableIdentity \cite{wang2024stableidentity} condition SD using an embedding basis derived from a celebrity dataset.
Deviating from CLIP features, methods such as Face0 \cite{valevski2023face0}, DreamIdentity \cite{chen2023dreamidentity}, IP-Adapter-FaceID \cite{ye2023ip} and PortraitBooth \cite{peng2023portraitbooth} employ robust ID-embeddings extracted from pre-trained face recognition models for conditioning the generation process. InstantID \cite{wang2024instantid} builds upon \cite{ye2023ip} by incorporating IdentityNet for enhanced control over identity and expression using facial landmarks, while ID-Aligner~\cite{chen2024idaligner} proposes reward-based feedback learning to further improve identity consistency. Finally, approaches such as \cite{yan2023facestudio, cui2024idadapter} combine both CLIP and identity embeddings for refined subject control.

Notably, all these methods prioritize text-driven recontextualization, using ID embeddings only as supplementary guidance. In contrast, the recently proposed Arc2Face~\cite{paraperas2024arc2face} model exclusively utilizes identity embeddings, effectively transforming SD into an identity-consistent facial foundation model by training on the large-scale WebFace~\cite{zhu2021webface260m} database. Our carefully designed distillation method leverages the facial prior in Arc2Face to achieve 3D avatar generation for any subject with superior realism and ID fidelity compared to existing techniques.

\subsection{3D Facial Generation}
\label{sec:related_animation}

Lifting the facial generation to 3D introduces several challenges.
The seminal approach of 3D Morphable Models (3DMM) 
\cite{blanz2003face, paysan20093d, li2017learning, egger20203d, smith2020morphable}
has long been used for facial shape and appearance modeling.
Generative models can be used on top of 3DMMs to model realistic skin appearance as textures,
aiming at generation \cite{gecer2020synthesizing, lattas2023fitme, li2020learning},
``in-the-wild'' reconstruction \cite{gecer2019ganfit, lattas2021avatarme++, luo2021normalized},
and inverse rendering \cite{dib2024mosar}; however, all are restricted to the skin region only.
Advances in Neural Radiance Fields (NeRF) \cite{mildenhall2021nerf}
have enabled recent works to represent the whole human head in high detail.
The use of tri-planes \cite{chan2022efficient} provides an efficient way for generative models 
to learn the distribution of human heads, with a GAN representation that is invertible for ``in-the-wild'' images.
PanoHead \cite{PanoHead} extends \cite{chan2022efficient} to tri-grids, enabling full 360° generation, and
Rodin \cite{wang2023rodin, zhang2024rodinhd} trains similar representations with GAN and diffusion backbones, using only synthetic data.
Implicit representations have also been used in conjuction with parametric models,
with MofaNeRF \cite{zhuang2022mofanerf}, HeadNeRF \cite{hong2022headnerf} and NeRFace \cite{gafni2021dynamic} achieving impressive rendering quality.
Similar volumetric priors can also be trained on synthetic data \cite{buhler2023preface} 
and have been shown to generalize to ``in-the-wild'' images.

Diffusion Models \cite{rombach2022high} and Score Distillation Sampling \cite{ruiz2023dreambooth} have recently been used to optimize facial avatars based on 2D priors. 
Magic123 \cite{Magic123} builds upon an implicit DMTET \cite{shen2021deep} representation to generate shapes with both 2D and 3D priors. DreamCraft3D \cite{sun2023dreamcraft3d} introduces a hierarchical pipeline for 3D mesh generation, 
though with limited success in facial geometry.
TADA \cite{liao2023tada} uses morphable model templates with SDS for text-to-3D generation, but suffers from highly saturated results for human faces.
Arguably, HumanNorm \cite{huang2023humannorm} first achieved crisp and detailed human avatar generation from text, using SDS, by relying on a normal generation model. Meanwhile, significant progress has been made in overcoming the limited quality of SDS results, by leveraging camera conditioning \cite{liu2023zero1to3, ReconFusion, ZeroNVS}, multi-view pre-training \cite{liu2023zero1to3, shi2023MVDream, wang2023imagedream, liu2023syncdreamer, gao2024cat3d} or improved distillation frameworks \cite{ProlificDreamer, ISM, tang2023dreamgaussian, shriram2024realmdreamer, miao2024dreamer}. However, all these works address generic 3D asset generation based on text or image inputs, which are insufficient to capture the nuanced details of facial identities. Closer to our work, ID-to-3D \cite{babiloni2024idto3d} extends the SDS approach with a DMTET representation for accurate shape generation and uses ID-embeddings instead of text to guide the optimization. However, it faces challenges in terms of the appearance quality due to the use of an optimizable UV texture.
In contrast, our work utilizes the recent breakthrough of 3DGS, and a 2D backbone specifically tailored to human faces to achieve human appearance reconstruction of superior quality.

\section{Method}
\label{sec:method}
\begin{figure*}[!ht]
    \centering
    \includegraphics[width=\linewidth]{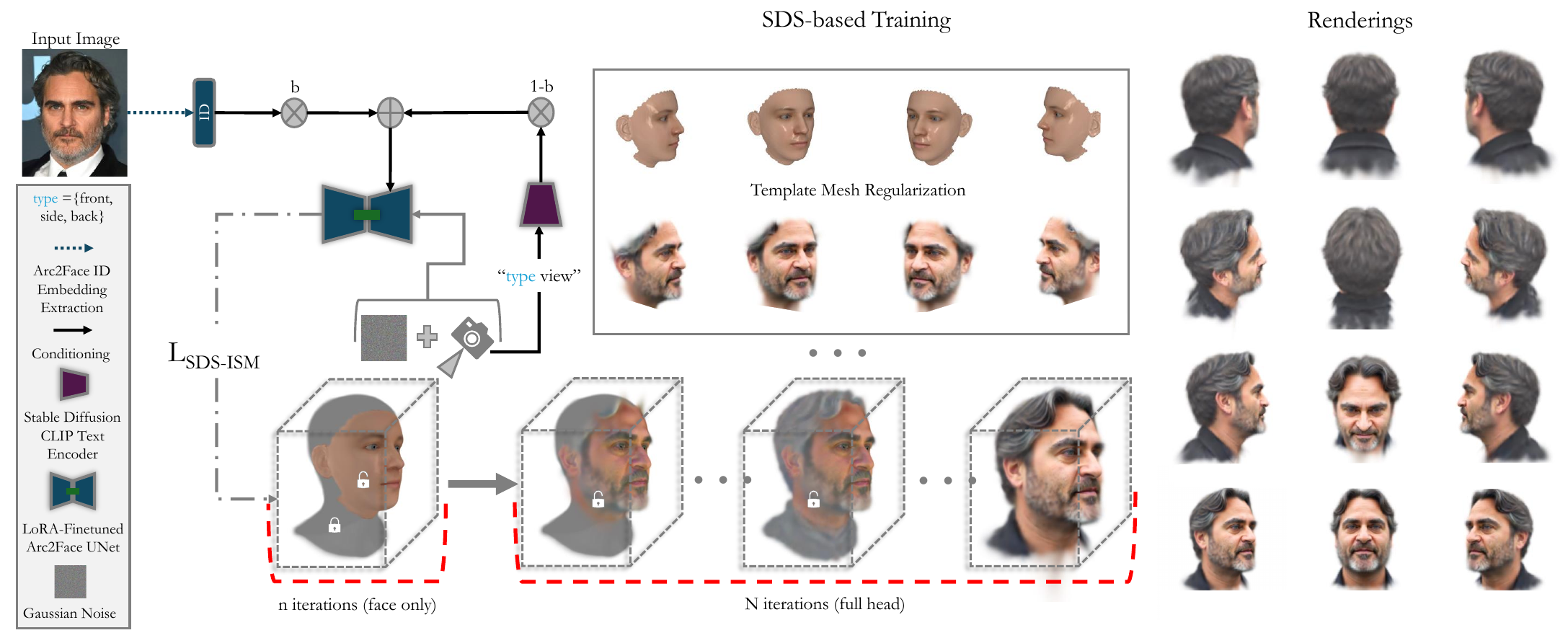}
    \caption{\textbf{Overview of the proposed 3D generation framework.} Our method uses an adapted Arc2Face diffusion model \cite{paraperas2024arc2face}, augmented for diverse view generation through fine-tuning on PanoHead \cite{PanoHead} samples. For 3D generation, starting with a frontal image, we extract the Arc2Face embedding and initialize Gaussian Splats on each vertex of the FLAME head model \cite{FLAME:SiggraphAsia2017}, fitting the facial area to the mean facial texture. We then apply an SDS alternative, where each iteration combines the Arc2Face embedding with a CLIP-encoded view embedding to denoise the renderings and update the splats. Initially, only facial splats are optimized for a set number of iterations. Subsequently, all splats are refined with densification, pruning, and opacity resets disabled for the facial area. Dense mesh correspondence is maintained through targeted initialization, avoidance of the standard 3DGS modifications in the facial region, and mesh regularizers adhering to the underlying template. Therefore, our method enables straightforward avatar expressions via blendshapes and shows potential for expression refinement after blendshape application using the same framework with minimal steps.}
    \label{fig:arc2avatar_framework}
\end{figure*}

Our method generates highly realistic and high-fidelity 3D head avatars by utilizing 3D Gaussian Splats (3DGS) \cite{kerbl3Dgaussians} and Score Distillation Sampling (SDS) \cite{poole2022dreamfusion}, guided by the robust Arc2Face \cite{paraperas2024arc2face} foundation model, using as few as a single facial image. To achieve this, we augment and fine-tune Arc2Face for diverse view generation. This results in a comprehensive model that generates diverse and ID-consistent head images spanning a wide (360°) range of viewing angles that serves as our guidance model. Additionally, our 3D avatars are expressive and can be easily deformed using FLAME \cite{FLAME:SiggraphAsia2017} blendshapes to convey a wide range of expressions. This is accomplished by modifying the splat optimization process and regularizing it to adhere closely to the underlying template and by effectively initializing the splats to enhance correspondence with the template. Finally, we include an SDS-based correction step that can refine the generated expressions. An illustration of our framework is provided in \cref{fig:arc2avatar_framework}. In the following, we first present the necessary background and then explain the components of our method in detail.

\subsection{Preliminary}
\label{sec:method_prelim}

\subsubsection{Score Distillation Sampling and Interval Score Matching}
\label{sec:method_prelim_sds}
SDS, introduced in \cite{poole2022dreamfusion}, has become the standard approach for text-guided 3D asset generation, serving as a bridge between 2D diffusion models and 3D representations. SDS operates by rendering views \( \mathbf{x_0} = f(\theta, \mathbf{c}) \) from a 3D representation parameterized by \( \mathbf{\theta} \). It then injects random noise into these renderings to obtain noisy latents \( \mathbf{x_t} \), for which a pretrained Denoising Diffusion Probabilistic Model (DDPM) \cite{ddpm_Ho} predicts the added noise \( \mathbf{\epsilon_{\phi}}(\mathbf{x_t}, t, \mathbf{y}) \), conditioned on text \( \mathbf{y} \). The SDS loss direction is defined by the following gradient:
\begin{align}
\nabla_{\mathbf{\theta}}\mathcal{L}_{\text{SDS}}(\theta) = \mathbb{E}_{t, \mathbf{c}} \left[ w(t) \left\| \mathbf{\epsilon_{\phi}}(\mathbf{x_t}, t, \mathbf{y}) - \mathbf{\epsilon} \right\|^2 \frac{\partial f(\mathbf{\theta}, \mathbf{c})}{\partial \mathbf{\theta}} \right],
\end{align}
where \( w(t) \) is a weighting function, \( \mathbf{\epsilon} \) is the actual noise added, and \( \mathbf{\epsilon_{\phi}}(\mathbf{x_t}, t, \mathbf{y}) \) is the predicted noise by the DDPM. As observed in \cite{ISM}, the SDS loss aims to align the rendered view \( \mathbf{x_0} \) with the pseudo-ground-truth (pseudo-GT) \( \mathbf{\hat{x}_t^0} \) estimated by the DDPM in a single-step prediction. However, SDS has two main weaknesses: pseudo-GTs \( \mathbf{\hat{x}_t^0} \) are often inconsistent due to DDPM sensitivity to noise and camera variations, and the one-step denoising leads to over-smoothed assets. 

To mitigate these, we opt to use Interval Score Matching (ISM) \cite{ISM}, which improves the pseudo-GT quality through DDIM inversion \cite{ddim}. Instead of generating \( \mathbf{x_t} \) stochastically, ISM employs DDIM inversion to create an invertible noisy latent trajectory, ensuring that the pseudo-GTs \( \mathbf{\tilde{x}_t^0} \) are consistently aligned with the original view \( \mathbf{x_0} \). Therefore, the loss direction for ISM is given by:
\begin{align}
\nabla_{\mathbf{\theta}} \mathcal{L}_{\text{ISM}}(\mathbf{\theta}) = 
\mathbb{E}_{t, \mathbf{c}} \Big[ & w(t) \left\| \mathbf{\epsilon_{\phi}}(\mathbf{x_t}, t, \mathbf{y}) - \mathbf{\epsilon_{\phi}}(\mathbf{x_s}, s, \varnothing) \right\|^2 \nonumber \\
& \times \frac{\partial f(\mathbf{\theta}, \mathbf{c})}{\partial \mathbf{\theta}} \Big],
\end{align}
where \( \mathbf{x_s} \) is an intermediate step obtained through DDIM inversion. Additionally, ISM employs multi-step denoising to produce higher-quality pseudo-GTs while maintaining computational efficiency, as it requires fewer iterations to converge compared to SDS. These two properties make ISM a superior approach compared to SDS, particularly for our task of 3D avatar generation, where consistency, symmetry, and high quality are crucial.

\subsubsection{Arc2Face}
\label{sec:method_prelim_arc2face}
Arc2Face \cite{paraperas2024arc2face} is a recently proposed powerful foundation model for human faces, leveraging a pre-trained Stable Diffusion (SD) \cite{rombach2022high} framework and an ArcFace \cite{deng2019arcface} feature extractor to generate identity-consistent facial images at a resolution of $512\times512$, from as little as a single input image of a subject. For a given facial image \( \mathbf{x} \in \mathbb{R}^{H \times W \times C} \), after alignment and cropping to the facial area, the ArcFace network is used to extract an identity embedding \( \mathbf{v} = \alpha(\mathbf{x}) \in \mathbb{R}^{512} \). To condition the SD model on this identity, the embedding must be adapted to SD conditioning standards. Arc2Face achieves this by projecting \( \mathbf{v} \) into the CLIP embedding space \cite{radford2021learning} using the original model's encoder \( \mathbf{E_\tau} \), which is fine-tuned to align with ArcFace embedding conditioning. Specifically, \( \mathbf{v} \) is zero-padded to form \( \hat{\mathbf{v}} \in \mathbb{R}^{768} \) and replaces a placeholder token in a fixed text prompt, creating a sequence of token embeddings \( s = \{e_1, e_2, \ldots, \hat{\mathbf{v}}, \ldots, e_N\} \). The encoder \( \mathbf{E_\tau} \) then maps this sequence to the CLIP latent space \( \mathbf{c} =  \mathbf{E_\tau}(s) \in \mathbb{R}^{N \times 768} \), which conventionally guides the U-Net through cross-attention layers to generate images that accurately reflect the input identity. This model is trained on the largest public face recognition dataset \cite{zhu2021webface260m} and is, thus, well-suited for our task, as its strong identity preservation ensures detailed, faithful, and consistent representations of the input subject.

\subsection{Extending Arc2Face for Diverse View Generation}
\label{sec:method_multiviewarc2face}
Standard models employed for SDS tasks are typically text-to-image models trained on diverse datasets, enabling them to generate various perspectives of objects. Arc2Face, however, is optimized to `overfit' on generating frontal and slightly angled facial images, making it unsuitable for wide view generation like other distillation models. This constraint is expected, as Arc2Face was not trained on back views and only minimally on side views. Therefore, we first extend Arc2Face for view-augmented generation, while preserving its strong frontal and slight side-view priors.

This requires the use of a multi-view face dataset for fine-tuning. Given that most existing ``in-the-wild'' facial datasets feature frontal images and collecting sufficient real-world data is challenging, we opt for a powerful 3D-aware generative model, namely PanoHead \cite{PanoHead}, to create a synthetic dataset of realistic multi-view head renderings. To minimize artifacts, especially apparent in PanoHead renderings of individuals with bald heads or long hair in non-frontal views, we curated 10,000 high-quality IDs from the initial set of generated heads. For each individual, we rendered approximately 26 images by rotating around the head over 360°, ensuring a diverse range of head perspectives.

Despite filtering, the domain gap between real-world data and PanoHead renderings can degrade the quality of generated images. While some degradation is mitigated by the averaging effect of SDS during the 3D optimization, minimizing such artifacts in the 2D backbone is essential. To achieve this, we use Low Rank Adaptation (LoRA) \cite{hu2022lora}, an efficient fine-tuning method that adapts Arc2Face by updating a reduced number of parameters in every attention-layer through low-rank approximation. After experimentation, we selected a rank of \( r = 16 \) for the LoRA matrices, achieving sufficient balance between novel-view adaptability and preserving Arc2Face's superior quality. 



\begin{figure}[h]
    \centering
    \includegraphics[width=0.5\textwidth]{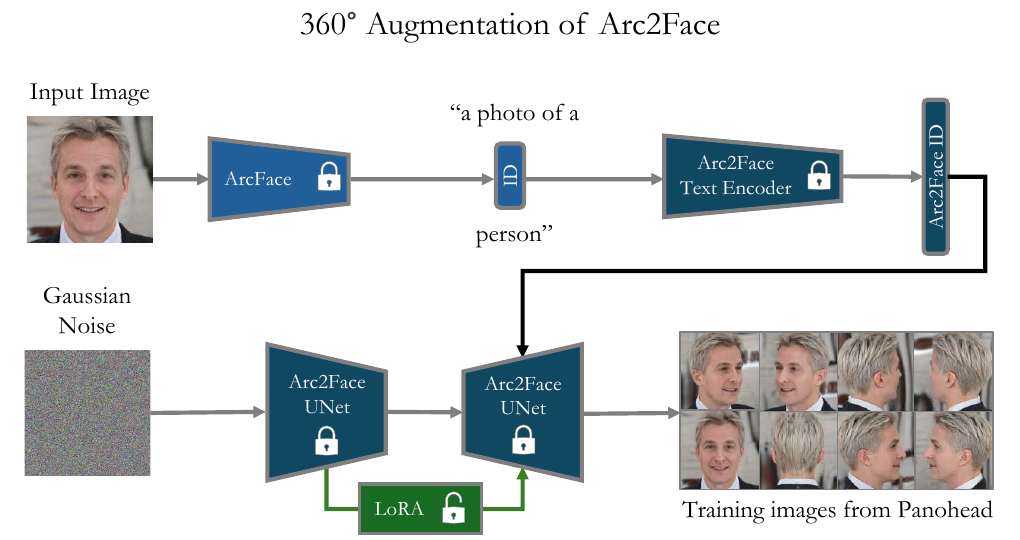}
    \caption{\textbf{LoRA \cite{hu2022lora} fine-tuning of Arc2Face \cite{paraperas2024arc2face} with PanoHead \cite{PanoHead} samples.} The generation is conditioned on the frontal sample.}
    \label{fig:method_panohead}
\end{figure}

During training, we extract the frontal image \( \mathbf{x_0^X} \)  of each individual \( \mathbf{X} \) to compute its ArcFace embedding \( \mathbf{v^X} = \alpha(\mathbf{x_0^X}) \). Each image in the set \( \{\mathbf{x_i^X}\}_{i=1}^{26} \) of an individual is conditioned on the same frontal \( \mathbf{v^X} \) when fed into the diffusion model for denoising, as depicted in \cref{fig:method_panohead}. The training loss for the denoising model \( \mathbf{\epsilon_{\theta, \lambda}}(\mathbf{x_{i, t}^X}, t, \mathbf{v^X}) \) is defined as:
\begin{align}
\mathcal{L}_{\text{LoRA}}(\mathbf{\lambda}) = \mathbb{E}_{t, \mathbf{\epsilon} \sim \mathcal{N}(0, \mathbf{I})} \left[ \left\| \epsilon - \mathbf{\epsilon_{\theta, \lambda}}(\mathbf{x_{i, t}^X}, t, \mathbf{v^X}) \right\|_2^2 \right],
\end{align}
with \( \mathbf{x_{i, t}^X} \) representing an image \( i \) of an individual \( X \) at time step \( t \), \( \mathbf{\theta} \) the fixed model weights, and \( \mathbf{\lambda} \) the trainable LoRA parameters,

\subsection{ID-guided 3D Head Generation using Arc2Face}
\label{sec:method_distil_arc2face}
Our distillation method builds upon the work of \cite{ISM}, which optimizes a 3D scene using ISM as an alternative to SDS. We extend this framework to reconstruct 3D head avatars from ID features using our view-augmented face model described above for guidance. 
During denoising, we carefully weight the added layers with an appropriate LoRA scale to allow generation of back and side views without compromising the quality of the generated images or degrading the identity. Specifically, through extensive experimentation, we selected a LoRA scale of \( s_{L} = 0.45 \), which provides a good balance.

Appropriate conditioning is essential for our model to differentiate between views while preserving the subject's identity. Regarding identity, we use the original Arc2Face conditioning, which involves an ArcFace embedding of the input subject. This embedding can either be derived from a single image \( \mathbf{v} \) or from averaging multiple embeddings \( \mathbf{v_i} \) from different photos \( \mathbf{v} = \frac{1}{N} \sum_{i=1}^{N} \mathbf{v_i} \). The ArcFace embedding is injected into the text embedding via a placeholder token \( \mathbf{id} \) in the default prompt ``photo of a \( \mathbf{id} \) person'' as described in \cite{paraperas2024arc2face}. Following that, we replace \( \mathbf{id} \) with \( \mathbf{v} \), resulting in an identity-conditioned text embedding:
\begin{align}
\mathbf{c_{\text{default}}} = \mathbf{E}_\tau(\text{"photo of a } \mathbf{id} \text{ person"}, \mathbf{v}),
\end{align}
where \( \mathbf{E_\tau} \) is the fine-tuned text encoder from Arc2Face. While Arc2Face has been overfitted on such identity-specific prompts, it remains fundamentally a SD model and can incorporate simple general textual guidance. Hence, to condition the model on different viewpoints, we modulate the identity-conditioned embedding with viewpoint-specific text embeddings obtained from the original SD text encoder \cite{radford2021learning} \( \mathbf{E_{\text{SD}}} \). We generate embeddings \( \mathbf{c_{\text{view}}} \) for simple directional prompts like ``front view'', ``side view'', and ``back view'' by passing them through \( \mathbf{E_{\text{SD}}} \):
\begin{align}
\mathbf{c}_{\text{view}} = \mathbf{E}_{\text{SD}}(\text{``\textbf{type} view''}),
\end{align}
to then create view-enriched embeddings by blending \( \mathbf{c}_{\text{default}} \) with \( \mathbf{c}_{\text{view}} \):
\begin{align}
\mathbf{c_d} = b \cdot \mathbf{c_{\text{default}}} + (1 - b) \cdot \mathbf{c_{\text{view}}},
\end{align}
where \( b \in [0, 1] \) balances the influence of identity and viewpoint, and \( d \) denotes the direction (``front'', ``side'', or ``back''). We set \( b = 0.85 \) to maintain a strong identity representation while incorporating viewpoint information.

\subsection{Mesh-Based Optimization}
\label{sec:method_mesh_optimization}
A commonly used approach for animating 3D avatars involves utilizing an underlying mesh with blendshapes. However, this approach poses challenges in an SDS-based 3DGS setup, as guiding the facial area's splats to adhere closely to an underlying template is challenging. We address this by employing a methodology that includes a masked 3DGS approach coupled with the FLAME \cite{FLAME:SiggraphAsia2017} 3D head model, the application of template correspondence regularizers and a strategic splat initialization.

First, we establish a dense correspondence between FLAME vertices and splats in the facial area, by assigning each template vertex a single splat, enabling very precise deformations with blendshapes. Densification, pruning, and opacity resets in facial splats disrupt this correspondence. While restricting these operations entirely would preserve the template, it would limit the modeling of complex areas like hair. Therefore, we adopt a masked 3DGS approach: we disable these operations only in the facial region, enabling them elsewhere for the best of both worlds.

To enhance correspondence with the underlying template, we incorporate two regularizers into the loss function for the masked splats. First, we employ the commonly used L2 regularizer to align the positions of the splats with their corresponding template vertices and prevent significant deviations from the template. Additionally, inspired by \cite{moon2024exavatar}, we use a connectivity-based regularizer, \ie the Laplacian, which is less common in 3DGS setups. Specifically, we minimize the difference between the Laplacian of the splat positions and the Laplacian of the corresponding vertices.

Prior to the SDS optimization, we begin with a densely upsampled version of the FLAME template and perform an initial number of iterations optimizing the splats \( G_{\text{init}}(\mathbf{x}) \) based only on the rendered mean face as guidance, i.e. using the following loss:
\begin{align}
\mathcal{L}_1 = \frac{1}{N} \sum_{i=1}^{N} \left| \mathbf{I_{\text{Gaussian}}}(i) - \mathbf{I_{\text{Mesh}}}(i) \right|
\end{align}
where \( I_{\text{Gaussian}} \) refers to the image generated by rendering only the splats that correspond to the facial region using the default Gaussian renderer, and \( I_{\text{Mesh}} \) denotes the image produced by rendering the points of \( G_{\text{init}}(\mathbf{x}) \) using a differentiable mesh renderer with the facial area's submesh connectivity and applying the standard mean texture from the FLAME template. The set of splats derived by this initial optimization closely resemble a mean facial surface, serving as a meaningful starting point for the susequent SDS process. At the same time, this approach ensures that \( G_{\text{init}}(\mathbf{x}) \) maintains high correspondence with the template mesh. 

We, then, perform our core SDS optimization process based on the Arc2Face denoising loss. In particular, we employ a few initial iterations focused on the masked facial area, using zoomed-in views to optimize only the corresponding splats, before initiating the standard optimization for all remaining splats. This 2-step SDS process is crucial, as experiments showed that using zoomed-out views from the start often disrupts facial texture alignment.

\subsection{Expression Generation}
\label{sec:method_expression_generation}
As described above, our strategy maintains a dense correspondence between facial splats and the FLAME template, enabling straightforward expression generation through precise blendshape deformations. While this approach typically produces realistic expressions, extreme expressions, such as opening the mouth, can result in missing or inaccurate splats. To address this, we employ our optimization to refine the avatar for blendshape-based expressions, enabling it to fill the mouth area with teeth and tongue. Initializing optimization from the deformed splat and performing only 500 iterations effectively corrects the expression. During this phase we do not use any regularizations to constrain the points to the template. This leads to expressions that are highly realistic and also diverse. Essentially, our framework allows quick refinements that preserve identity while enhancing realism.

\begin{figure*}[h!]
    \centering
    \includegraphics[width=\textwidth]{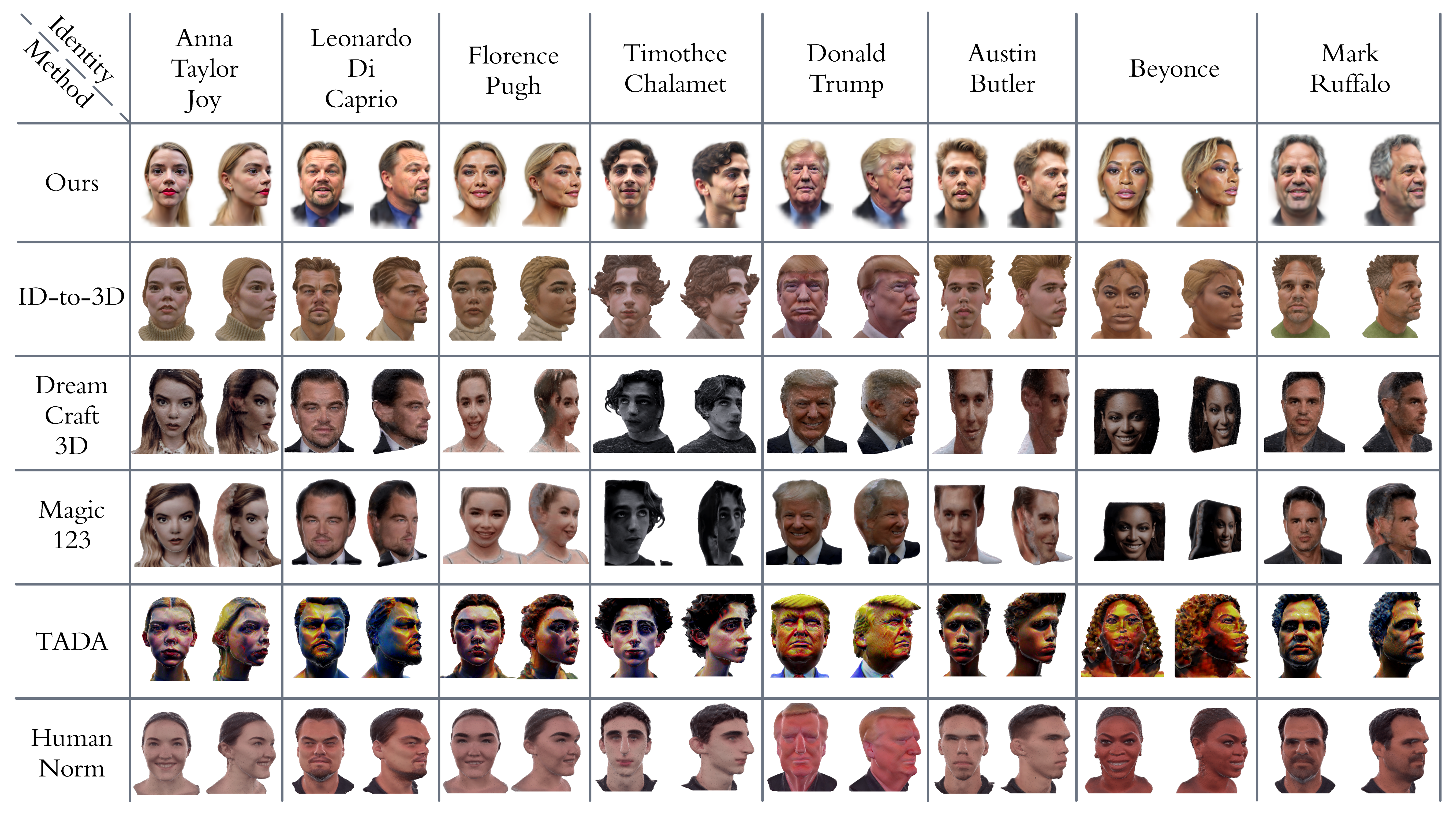}
    \caption{\textbf{Qualitative comparison of competing methods for celebrities.} ID-methods use an average ID-embedding from multiple images, although they can also use that of a single image.}
    \label{fig:qualitative_comparison}
\end{figure*}

\section{Experiments}
\label{sec:experiments}

In this section, we compare our method with state-of-the-art diffusion-based 3D avatar generation methods. Although our approach and our primary competitor \cite{babiloni2024idto3d} are the only identity-driven methods, we also compare against related text-to-3D \cite{huang2023humannorm, chen2023fantasia3d, liao2023tada, zhang2023dreamface} and image-to-3D methods \cite{Magic123, sun2023dreamcraft3d}, both quantitatively and qualitatively. Following \cite{babiloni2024idto3d}, we selected a set of 30 celebrities, where for each celebrity we used 20 images for evaluation and 5 images to compute an average identity embedding to condition the generation, although both our method and ID-to-3D \cite{babiloni2024idto3d} can work with a single image.

\subsection{Qualitative Comparison}
\label{sec:experiments_qualitative}
In \cref{fig:qualitative_comparison}, we showcase generated 3D celebrity identities from the methods that consistently produce the most meaningful representations for subjective evaluation. Specifically, while DreamFace \cite{zhang2023dreamface} and Fantasia3D \cite{chen2023fantasia3d} are included in our quantitative analysis, we exclude them from the qualitative comparison due to their limited ability to generate meaningful 3D identities and for space limitations.

It is evident that our method achieves superior realism while effectively preserving identity. We capture sufficient details with natural and accurate colors, notably avoiding the common oversaturation issues associated with SDS approaches, thereby highlighting the effectiveness of our carefully designed SDS strategy. Additionally, we successfully model challenging areas such as hair. In contrast, while ID-to-3D \cite{babiloni2024idto3d} produces sharp geometry, its representations can appear somewhat unrealistic and lack accuracy in certain aspects. Comparing our method with image-to-3D approaches \cite{Magic123, sun2023dreamcraft3d} based solely on frontal views may not provide a complete picture, as these methods primarily transfer 2D images to 3D space, maintaining most of their original appearance. Consequently, their side views often exhibit significant artifacts and reduced quality.

\subsection{Quantitative Comparison}
\label{sec:experiments_quantitative}
\begin{figure}
\centering
  \begin{minipage}[t]{0.55\linewidth}
    \centering
    \raisebox{-1.5ex}{
    \includegraphics[width=\linewidth]{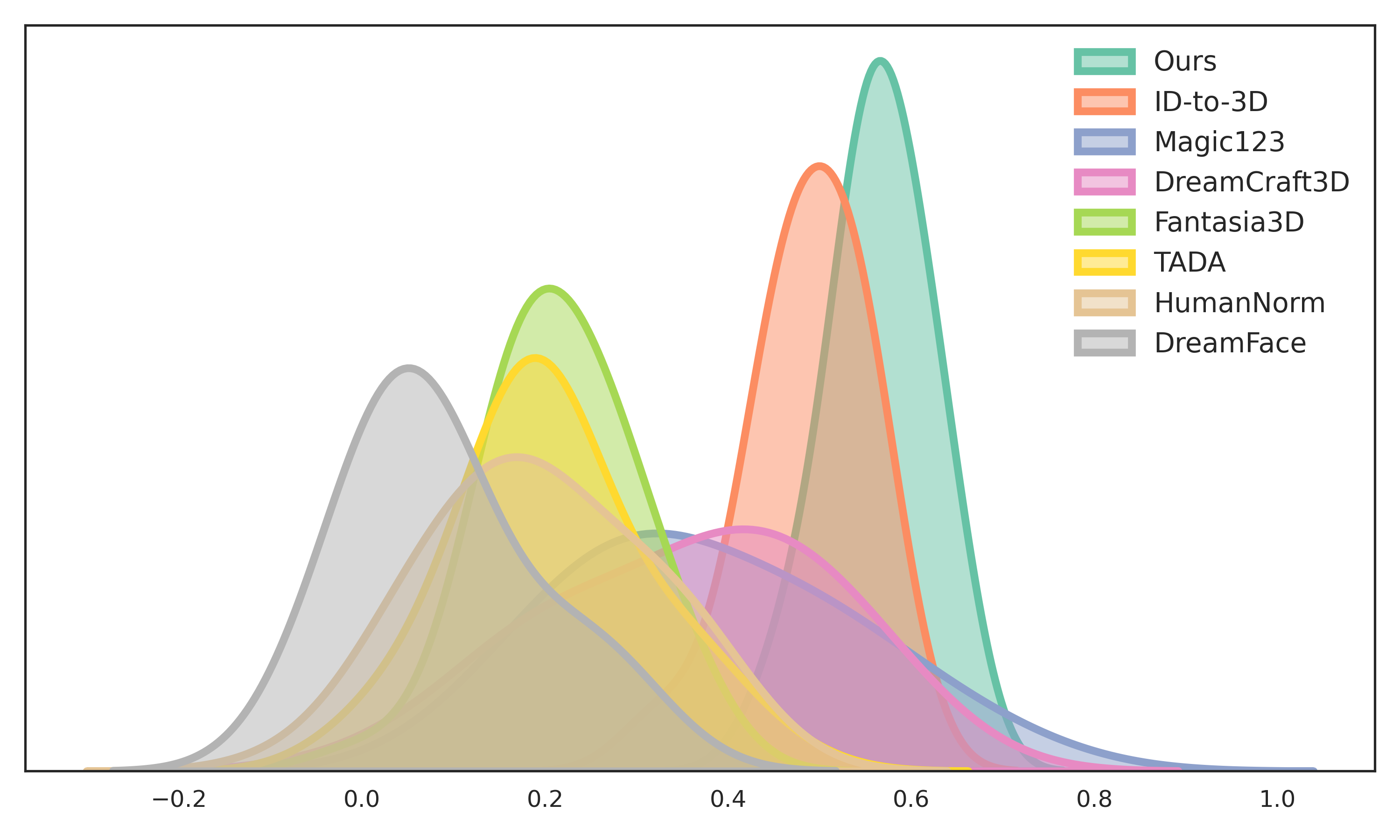}
    }
    \vspace{0.15cm}
    \captionof{figure}{\textbf{Identity similarity distribution for quantitative comparison.}}
    \label{fig:quantitative_comparison}
  \end{minipage}%
  \hfill
  \begin{minipage}[t]{0.4\linewidth}
    \centering
    \setlength{\tabcolsep}{1.8pt}
    \scriptsize
    \raisebox{9.5ex} {
    \begin{tabular}{lc} 
        \toprule
         & FID \\
         \midrule
         TADA \cite{liao2023tada} & 213.39\\
         DreamFace \cite{zhang2023dreamface} & 214.58\\
         Fantasia3D \cite{chen2023fantasia3d} & 280.32\\
         HumanNorm \cite{huang2023humannorm} & 173.02\\
         ID-to-3D \cite{babiloni2024idto3d} & 154.51\\
         Magic123 \cite{Magic123} & 159.21 \\
         DreamCraft3D \cite{sun2023dreamcraft3d} & 186.98\\
         \textbf{Ours} & \textbf{144.58}\\
        \bottomrule
    \end{tabular}
    }
    \captionof{table}{\textbf{Quantitative comparison based on FID.} }
    \label{tab:FID}
  \end{minipage}
\end{figure}

To quantitatively validate the superiority of our method, we utilize the standard Identity Similarity Distribution (ISD) benchmark, which effectively measures identity preservation and fidelity. We rendered 3D assets from all methods to obtain frontal and side views, as only our method and ID-to-3D are capable of generating realistic back views. Using ArcFace \cite{deng2019arcface}, we extracted identity embeddings from both the rendered images and the evaluation set for each celebrity. We then calculated the cosine similarity between the rendered embeddings and the real image embeddings to measure identity similarity. Grouping by method, we present the distribution of similarity scores for each approach in \cref{fig:quantitative_comparison}.


The distribution of similarity scores highlights the strength and robustness of our method, with ID-to-3D being the only other approach that consistently captures identity across different views. Our method achieves both the highest scores and the lowest variance, clearly outperforming the main competitor. Although image-to-3D methods can attain high scores, they also show high variance and some very low scores. This inconsistency arises because their frontal views closely mimic the input image, while other views contain significant artifacts. These results indicate that our avatars are characterized by the highest identity consistency and uniformity across views.

In order to additionally assess the realism of each method's avatars, we used the Fréchet Inception Distance (FID) metric \cite{heusel2017gans, Seitzer2020FID}, calculated between the renderings and the ground truth celebrity images. As shown in \cref{tab:FID}, our method achieves the lowest FID score, confirming its superior quality in the generated faces.


Finally, to further validate our findings, we conducted a user study with 40 participants. Each participant was shown renderings from the four best-performing methods, each applied to a random selection of 10 identities, and was asked to choose the most realistic and faithful avatar per identity. Notably, as shown in \cref{tab:experiments_userstudy}, our method received 93\% of the votes, indicating strong user preference and positive feedback regarding the quality of our avatars.

\subsection{Expression Evaluation}
\label{sec:experiments_expression}
\begin{figure}[h!]
    \centering
    \includegraphics[width=0.49\textwidth]{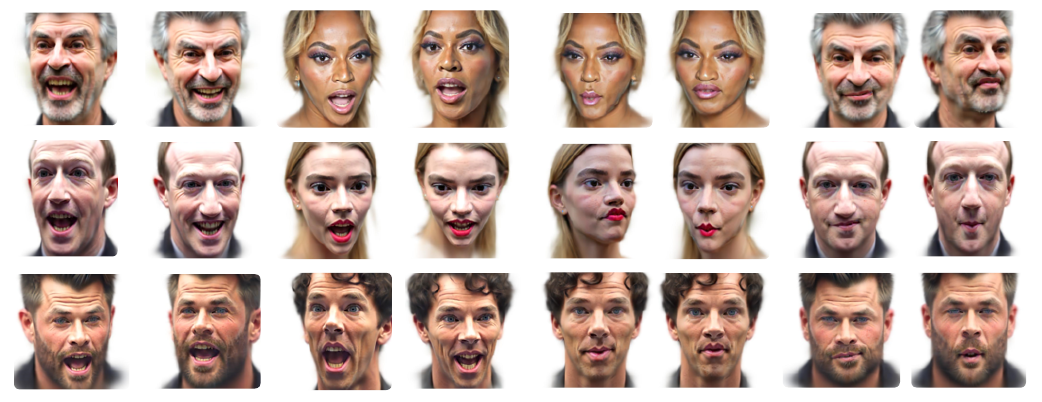}
    \caption{\textbf{Expression generations for subjective evaluation.} Open-mouth expressions (\ie columns 1,2,3,4) are corrected via our SDS refinement step.}
    \label{fig:experiments_exp_qualitative}
\end{figure}

In this subsection, we present visualizations to evaluate the fidelity and ID consistency of our generated expressions. First, we showcase the high fidelity of our expressions by applying a subset of challenging CoMA \cite{COMA} expression blendshapes. For extreme cases or open-mouth poses, we utilize our SDS correction step to refine the results. As shown in \cref{fig:experiments_exp_qualitative}, our method generates realistic expressions even for this challenging dataset. Furthermore, our refinement step succeeds in optimizing unoptimized mouth areas, accurately rendering tongue and teeth to convey open-mouth expressions with high fidelity. For additional qualitative results, please refer to the Supp.~Material.

\begin{figure}[h]
\centering
\begin{minipage}[t]{0.3\linewidth}
    \centering
    \setlength{\tabcolsep}{0.7pt}
    \scriptsize
    \raisebox{9.5ex} {
    \begin{tabular}{lc}
        \toprule
        & Pref.\\
        \midrule
        ID-to-3D \cite{babiloni2024idto3d} & 7\%\\
        Magic123 \cite{Magic123} & 0\%\\
        DreamCraft3D \cite{sun2023dreamcraft3d} & 0\%\\
        \textbf{Ours} & \textbf{93\%}\\
        \bottomrule
    \end{tabular}
    }
    \captionof{table}{\textbf{Preference user study results.} }
    \label{tab:experiments_userstudy}
  \end{minipage}
  \hfill
  \begin{minipage}[t]{0.67\linewidth}
    \centering
    \includegraphics[width=\linewidth]{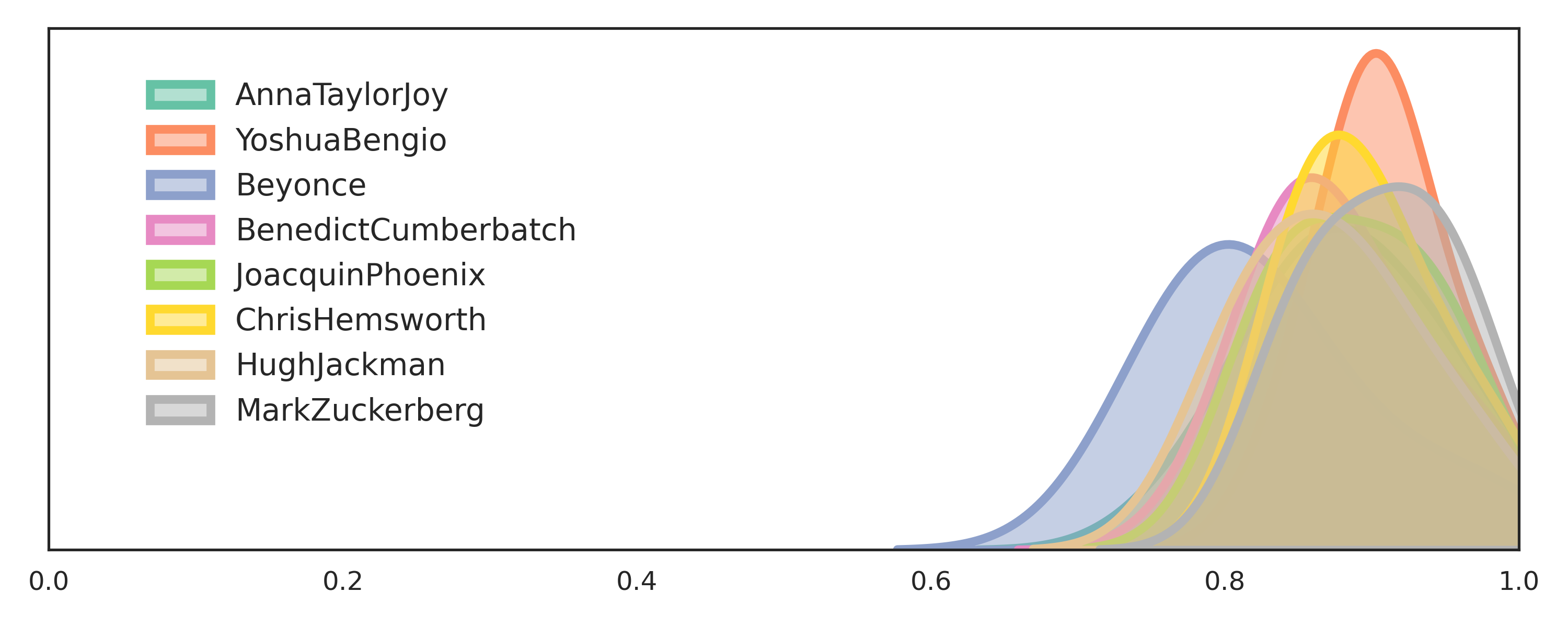}
    \captionof{figure}{\textbf{Identity Similarity Distribution between expressive and neutral renderings per celebrity.}}
    \label{fig:experiments_exp_idsim}
  \end{minipage}%
\end{figure}

Finally, to demonstrate the identity consistency of our generated expressions, we evaluated the identity similarity distribution by calculating the similarity between expressive renders and their neutral counterparts, following the approach of \cite{babiloni2024idto3d}. In \cref{fig:experiments_exp_idsim}, we present the distribution for each celebrity. As shown, our generated expressions remain consistent with high similarity scores and little variance.

\section{Limitations and Future Work}
\label{sec:limit_futur}
While our method effectively generates expressive 3D avatars, there are some areas that could be improved. Firstly, although our initialization strategy and regularizers typically enforce a neutral pose, there are occasional instances where this neutrality is lost due to the stochasticity in facial expression introduced by Arc2Face, resulting in misaligned correspondence. Future work could address this by integrating an expression-conditioned extension of Arc2Face to ensure consistent neutrality during optimization. Additionally, artifacts sometimes appear in the ears, which can be attributed to limitations of PanoHead \cite{PanoHead}. Exploring alternatives, such as the recently proposed SphereHead method \cite{li2024sphereheadstable3dfullhead}, may help reduce these errors. Furthermore, similar to many distillation techniques, our splats can become blurry near the edges. This issue could potentially be mitigated in future iterations through more strategic management of densification, opacity, and pruning processes, or by using a higher-resolution diffusion backbone.\\
\textbf{Impact.} Importantly, the ability to generate highly realistic 3D avatars of any individual introduces significant ethical concerns, including the risk of misuse for creating deceptive deepfakes. We strongly believe it is crucial to implement this technology responsibly, respecting individual privacy, and preventing unauthorized or harmful applications.

\section{Conclusion}
\label{sec:conclusion}
We introduced Arc2Avatar, an SDS-based 3D face generation approach that combines 3D Gaussian Splats with an ID-driven 2D diffusion model to achieve avatars with unmatched similarity and realism compared to existing approaches. Through lightweight fine-tuning on synthetic head images, we augment a state-of-the-art face model, Arc2Face, for side and back-view generation, while preserving ID consistency. Our distillation framework incorporates strategic initialization and mesh-based regularization, producing 3D avatars that retain dense correspondence with the underlying 3DMM, and can be driven by blendshapes. Additionally, our expression-specific correction step further refines accuracy. Our comparisons against related methods demonstrate the advantages of our pipeline, which offers an efficient solution for controllable, realistic 3D avatars. 

\section*{Acknowledgements} S. Zafeiriou and part of the research were funded by EPSRC Fellowship DEFORM (EP/S010203/1) and EPSRC Project GNOMON (EP/X011364/1). R.A. Potamias was supported by ESPRC Project GNOMON (EP/X011364/1). Finally, we would like to thank Francesca Babiloni and George Kopanas for their valuable comments and feedback on this work.

{
    \small
    \bibliographystyle{ieeenat_fullname}
    \bibliography{main}
}


\clearpage

\appendix  

\setcounter{section}{0}
\renewcommand{\thesection}{\Alph{section}}  

\twocolumn[{
\centering
\vspace*{1.5em}  
{\Large \bf Arc2Avatar: Generating Expressive 3D Avatars \\
from a Single Image via ID Guidance \\
(Supplementary Material)\par}
\vspace*{1.5em}
}]

This document offers supplementary details about our method that could not be included in the main paper due to space constraints. Additionally, it presents further qualitative results and includes an accompanying video that provides an overview of the proposed approach.

\section{Optimization of the Facial Area}
In this section, we illustrate the avatar optimization process for the facial area, highlighting the efficiency of our framework in swiftly capturing the person's facial features through strategic initialization, before advancing to the optimization of the entire head.

\subsection{Fitting Splats to the Mean Facial Surface}
\begin{figure}[h]
    \centering
    \includegraphics[width=\linewidth]{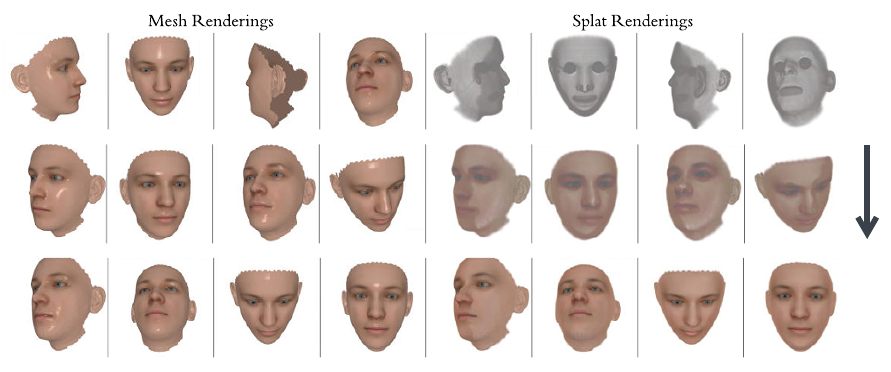}
    \caption{\textbf{Initial mean texture fitting.} The splats in the facial area are optimized based on mesh renderings with the mean texture. In the end, the splats closely replicate the mean textured mesh.}
    \label{fig:mean_init}
\end{figure}

As discussed in the main paper (Sec.~3.4), it is beneficial to initialize person-specific avatar generation using a set of Gaussian Splats representing the mean colored facial surface, rather than just the upsampled FLAME point cloud. This approach requires an initial optimization of all splats' parameters (including covariance) based on mesh renderings with the mean texture, as merely assigning the RGB values of each vertex to the point cloud would still result in a discontinuous representation. This fitting process is described in Sec.~3.4 and is further illustrated in \cref{fig:mean_init} for clarity. Please note that this step is independent of the input subject and is performed only once. The fitted Gaussian Splats, being identical for all subjects, serve as a precomputed initialization for subsequent subject-specific optimizations. 

\begin{figure}[h]
    \centering
    \includegraphics[width=\linewidth]{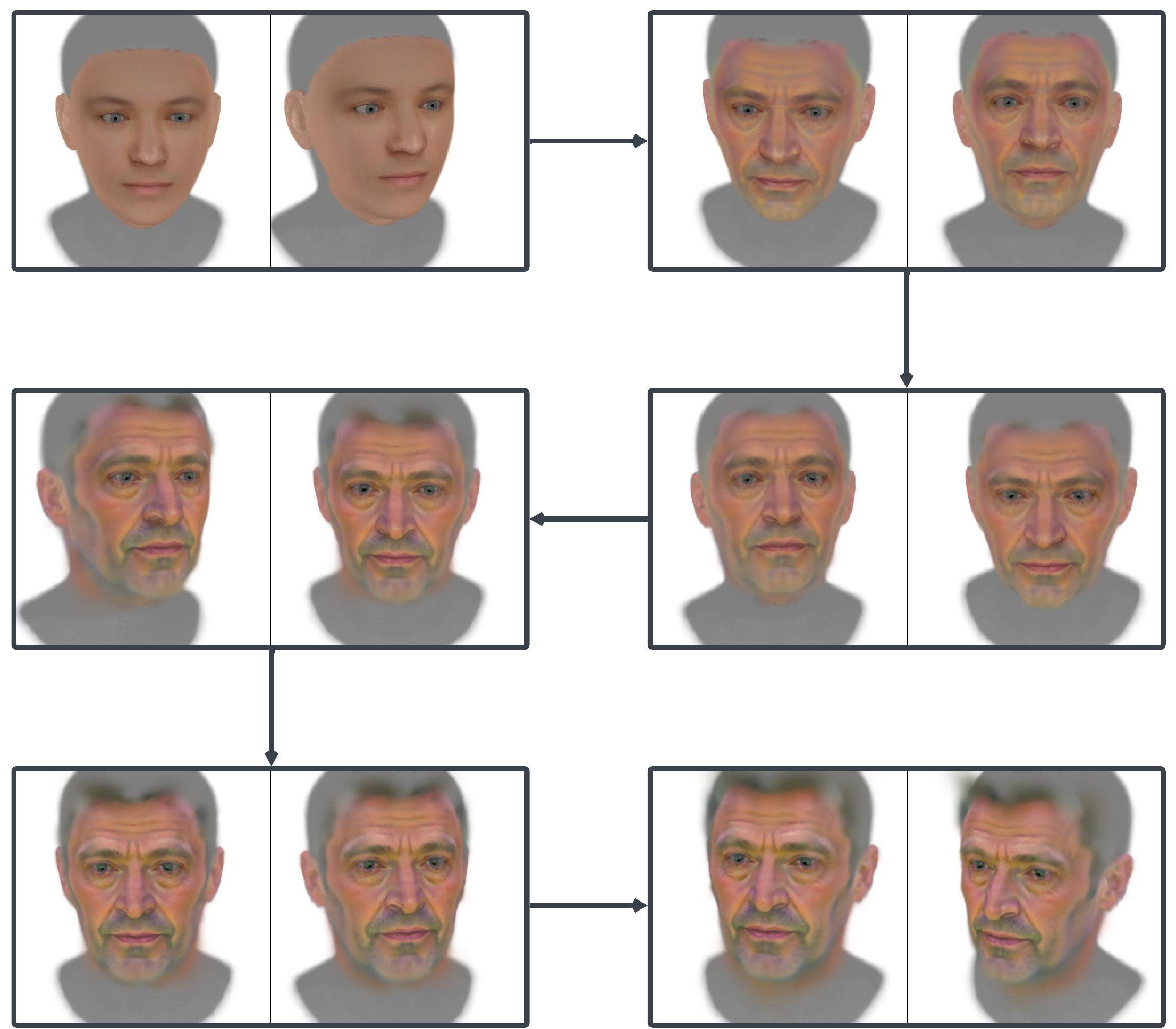}
    \caption{\textbf{Face-only optimization progression.} The splats in the facial area smoothly transition from the mean texture to the input subject's face.}
    \label{fig:init_prog}
\end{figure}

\subsection{Personalization of the Facial Area}
\label{sec:implement}
As described in the paper, our person-specific SDS optimization begins with the mean texture-fitted splat and proceeds in two phases: first optimizing the facial region, followed by optimizing the entire head.
The initial face-only phase consists of \(500\) iterations, during which we sample views with azimuth angles in the range \([-110, 110]\) and elevation angles in the range \([60, 90]\). To capture finer details, we zoom in during this phase, using a field of view of \(0.4\). In \cref{fig:init_prog}, we illustrate the progression of this initial optimization, showing how the mean texture-fitted splat smoothly transforms into the subject's face while maintaining correspondence with the underlying template mesh, highlighting the effectiveness of our approach.

\section{Ablation Studies}
Below, we provide additional results demonstrating the necessity of various components in our pipeline.

\subsection{Importance of Mean Texture Initialization}
\begin{figure}[h]
    \centering
    \includegraphics[width=\linewidth]{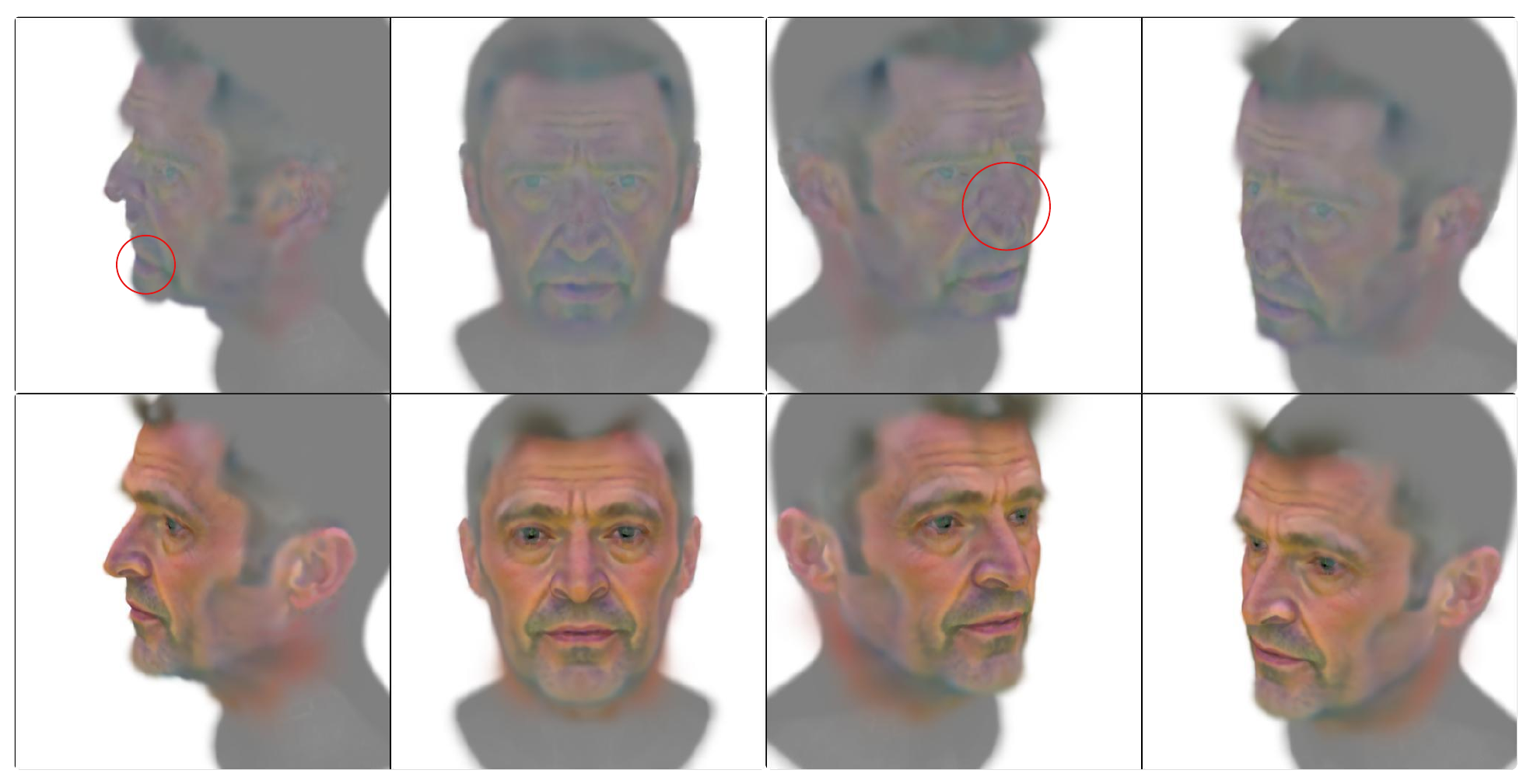}
    \caption{\textbf{Mean texture initialization impact.} Although template regularization achieves geometrical correspondence, the absence of the proposed initialization (top) leads to significant texture misalignment, disrupting the overall correspondence. Without mean texture initialization, facial features are often placed to incorrect locations (e.g., the mouth placed on the chin) or exhibit artifacts, such as duplicate features (e.g., the nose). In contrast, mean texture initialization (bottom) ensures proper correspondence from the early stages of the process.}
    \label{fig:no_text}
\end{figure}

We argue that template regularizers alone are insufficient to guarantee template correspondence in an SDS setup. To support this claim, we present results from the initial facial area optimization phase, comparing scenarios with and without mean texture initialization, as shown in \cref{fig:no_text}.

\subsection{Arc2Face Augmentation and View Embeddings}
Furthermore, we  demonstrate the necessity of both LoRA-based Arc2Face fine-tuning and the use of view text embeddings in conjunction with the identity embedding for conditioning. These processes are crucial for achieving realistic 3D avatars without Janus artifacts or inconsistencies.

As described in the main paper, we create view-enriched embeddings by blending the default identity embedding \( \mathbf{c}_{\text{default}} \) with the view embedding \( \mathbf{c}_{\text{view}} \) using the formula:
\begin{align}
\mathbf{c}_d = b \cdot \mathbf{c}_{\text{default}} + (1 - b) \cdot \mathbf{c}_{\text{view}},
\end{align}
where \( b \in [0, 1] \) balances the influence of identity and view. 

\begin{figure}[h]
    \centering
    \includegraphics[width=\linewidth]{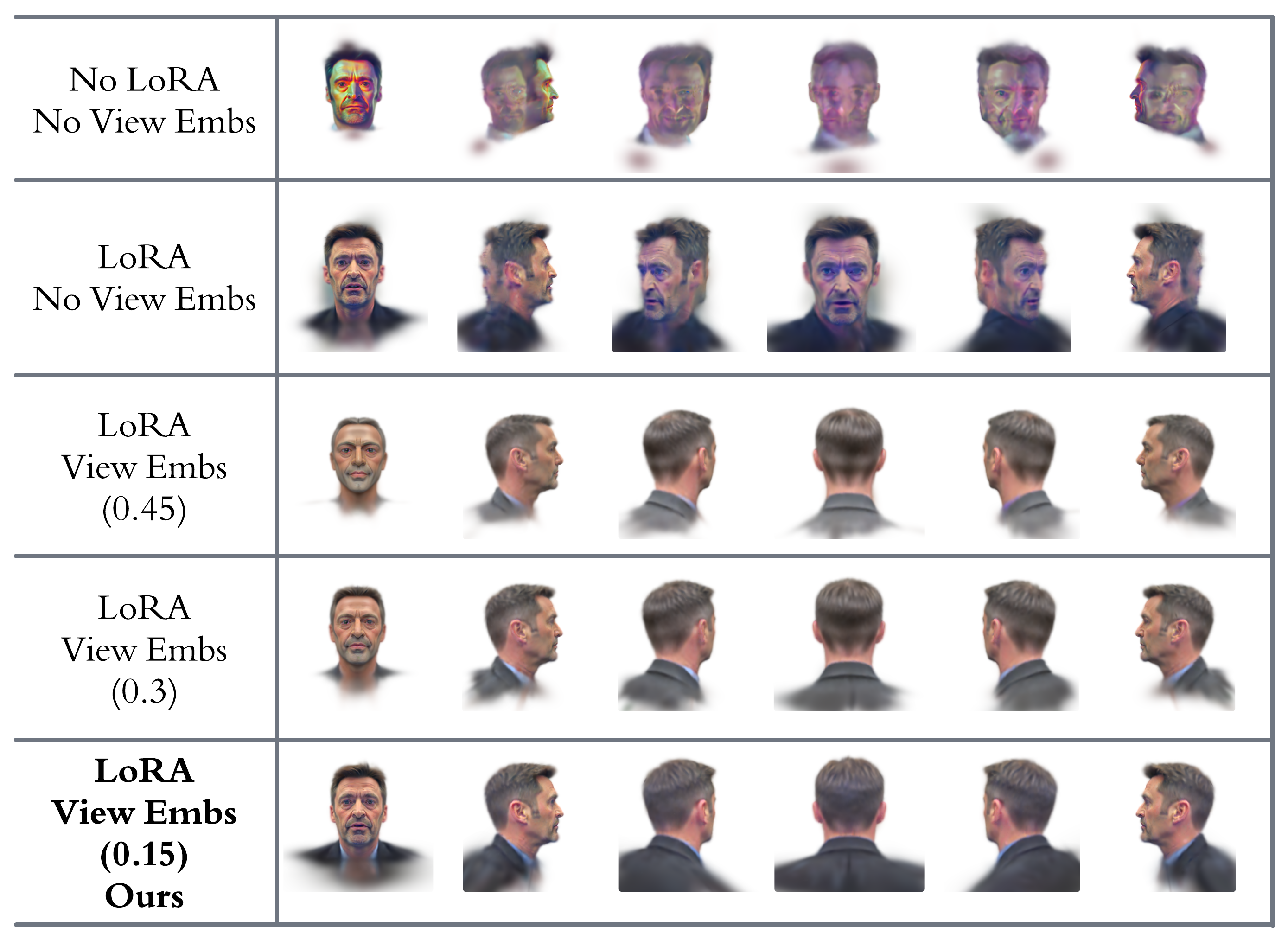}
    \caption{\textbf{Impact of augmenting Arc2Face for 360° generation and using view embeddings during distillation.} As expected, the default model is limited to modeling the frontal view, resulting in multiple inconsistencies and Janus artifacts in other views, as well as oversaturated colors, rendering it unsuitable for guidance in its default state. The LoRA-extended model without view embeddings performs better, achieving good identity preservation and improved side views, but it still exhibits Janus effects in the back view. Using strong weights for view embeddings generates very consistent heads with good back and side views but significantly reduces identity fidelity. In contrast, our selection of a low weight for view embeddings (final row) achieves the best of both worlds, combining identity preservation with consistency in the generated heads, and eliminating Janus effects.}
    \label{fig:ablation_study}
\end{figure}

In \cref{fig:ablation_study}, we present renderings of the avatar produced after the first half of the optimization steps for five different variations of our method:

\begin{enumerate}
    \item \textbf{Default Arc2Face Model:} Using the default ID-conditioned Arc2Face model as the guidance model without any modifications.
    \item \textbf{LoRA-Extended Model without View Embeddings:} Using the LoRA-extended model but without view embeddings, effectively setting the view embedding weight to zero (\( 1 - b = 0 \)).
    \item \textbf{Strong View Embedding Weight (\( 1 - b = 0.45 \)):} Using a strong weight for the view embedding to emphasize view information.
    \item \textbf{Medium View Embedding Weight (\( 1 - b = 0.3 \)):} Using a medium weight for the view embedding, providing a balanced influence between identity and view.
    \item \textbf{Our Method (\( 1 - b = 0.15 \)):} Using the blending factor we chose for our method, which we found to offer the best trade-off between identity preservation and view consistency.
\end{enumerate}

\begin{figure*}[t]
    \centering
    \includegraphics[width=\linewidth]{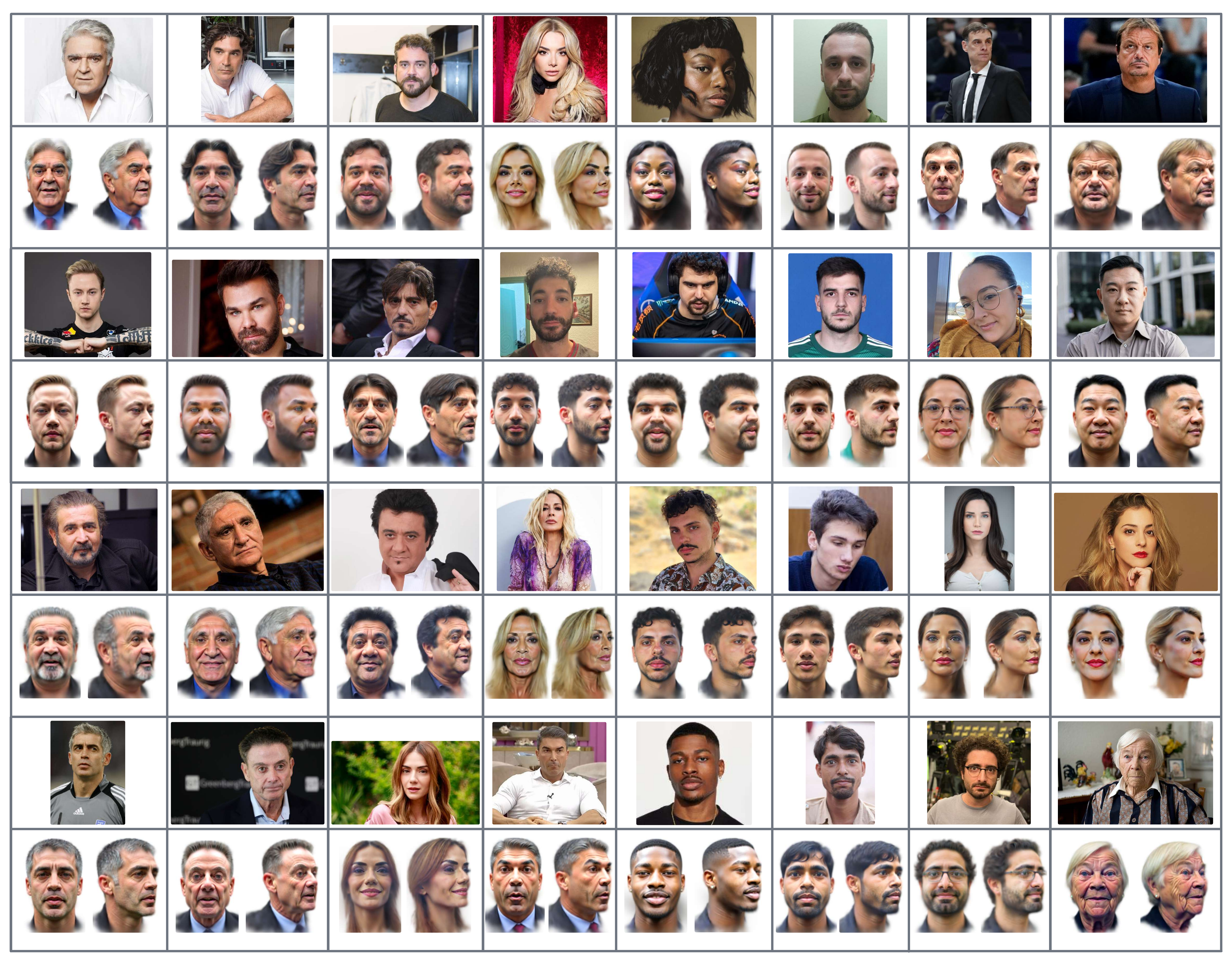}
    \caption{\textbf{Arc2Avatar is not limited to celebrities.} 
    Our method exhibits strong generalization, providing realistic and consistent 3D avatars for individuals of different ages, ethnicities, and backgrounds.
    }
    \label{fig:non_celebrities}
\end{figure*}

\section{Additional Qualitative Results}

In this section, we showcase additional 3D avatars generated by our method for subjects with significantly diverse characteristics. 
As can be seen in \cref{fig:non_celebrities}, our method exhibits strong generalizability, capable of producing high-fidelity, ID-consistent 3D heads for any individual.

\begin{figure*}[t]
    \centering
    \includegraphics[width=0.9\linewidth]{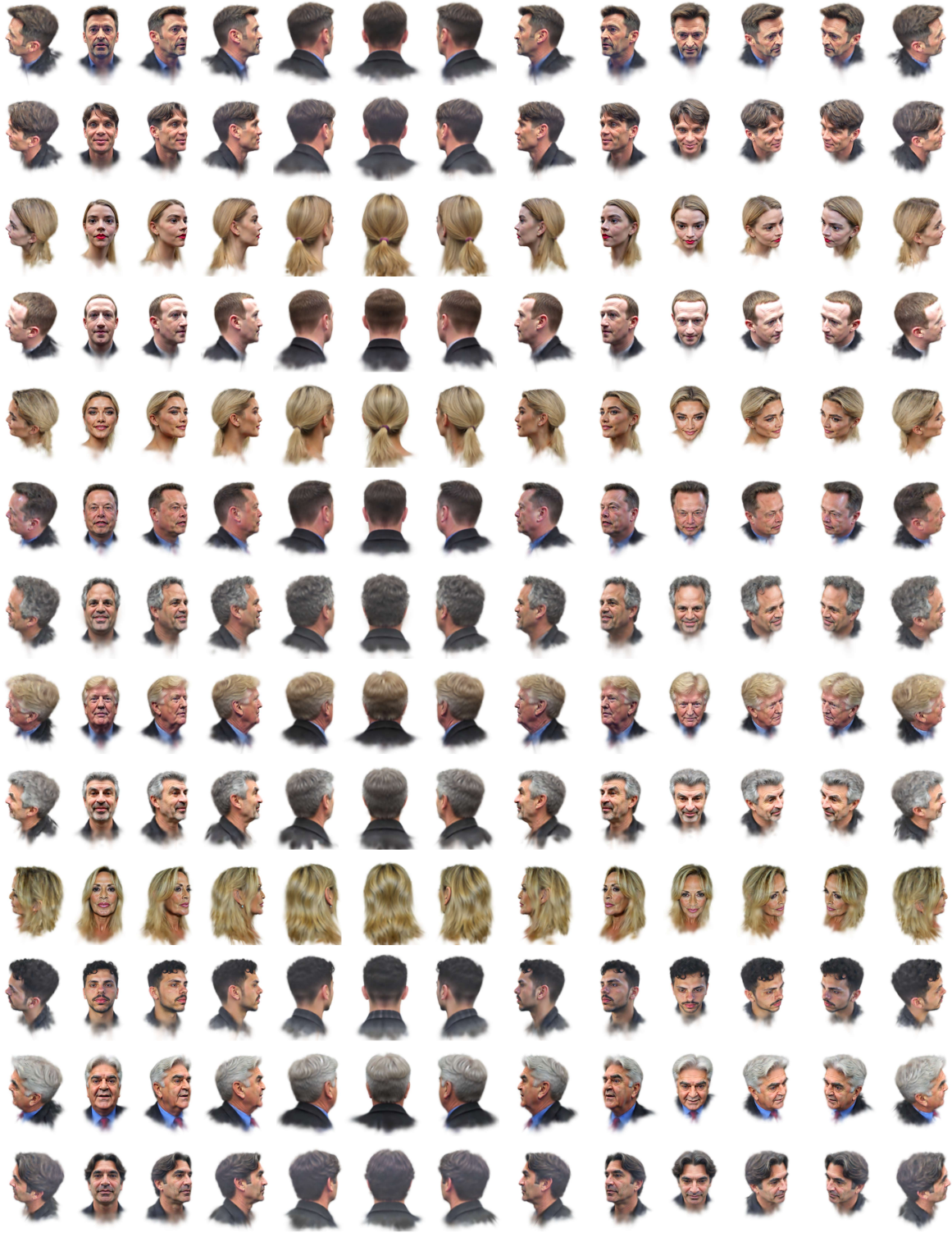}
    \caption{\textbf{Renderings of generated 3D avatars from diverse viewpoints.} Our method extends beyond realistic frontal views to produce complete 3D head models that can be rendered from any angle.}
    \label{fig:multiview}
    \vspace{-0.3cm}
\end{figure*}

Moreover, in \cref{fig:multiview}, we provide renderings from multiple perspectives for many samples, demonstrating our method’s 3D consistency and fidelity. Notably, our approach effectively generates realistic views, including challenging back-head perspectives, which are inferred solely from frontal input images thanks to our careful adaptation of Arc2Face for diverse view generation using frontal inputs.

\section{Failure Cases}
\begin{figure}[t]
    \centering
    \includegraphics[width=\linewidth]{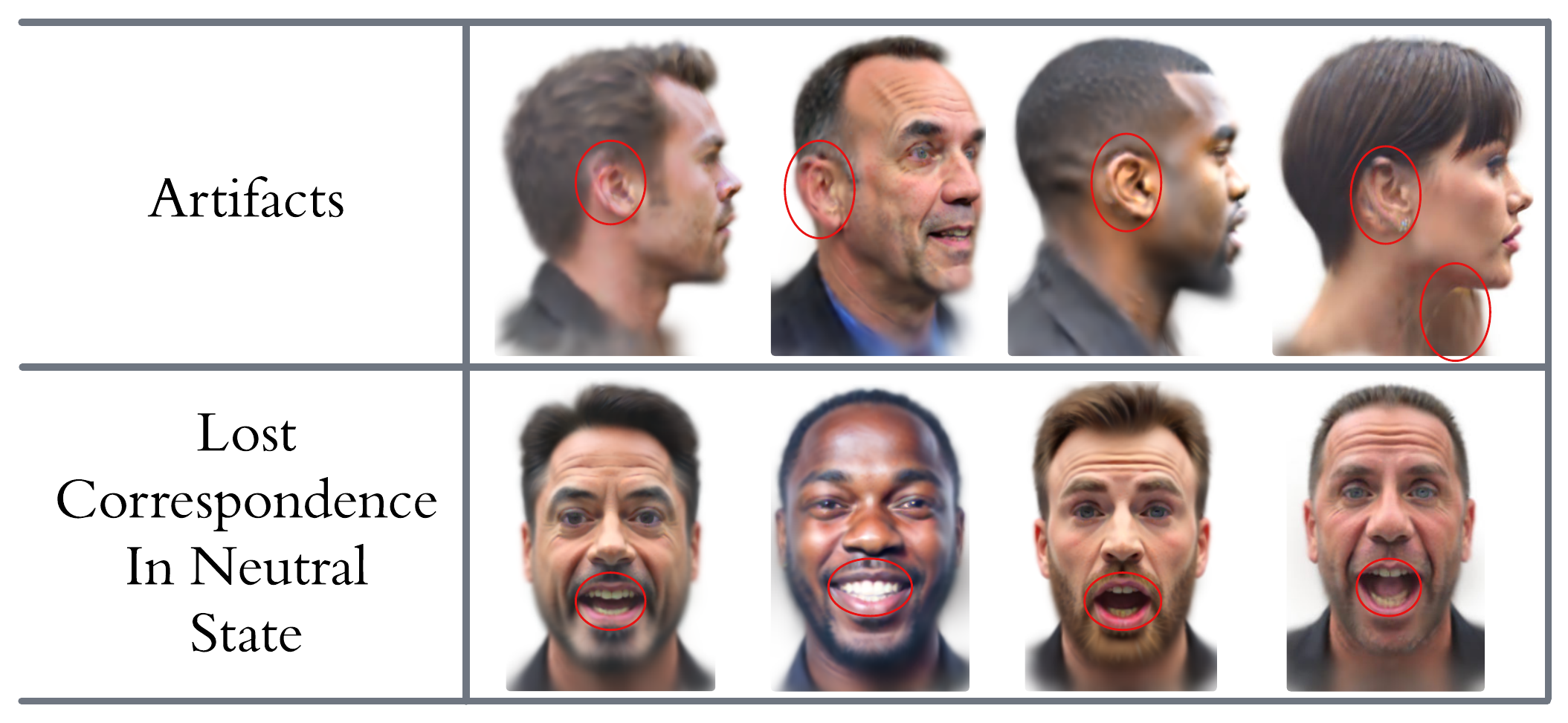}
    \caption{\textbf{Failure cases.} Artifacts may appear around the ears and neck regions. Additionally, certain inputs can bias the optimization towards smiling or surprised expressions, despite the underlying neutral mesh. Nevertheless, the avatars consistently preserve the individuals' identities.}
    \label{fig:failure}
\end{figure}

As discussed in the main paper, our method has certain limitations, including the introduction of artifacts and the occasional addition of expressions by Arc2Face in the neutral optimization stage despite our efforts to enforce consistency with the neutral mesh, disrupting correspondence with the template. In \cref{fig:failure}, we present examples showing these issues.


\section{Implementation Details}
\label{sec:implement}

\subsection{Arc2Face Fine-Tuning}
\label{sec:implement_ext}

We fine-tuned the LoRA-augmented Arc2Face model following a setting similar to \cite{paraperas2024arc2face}. In particular, we used a resolution of $512\times512$ pixels for our synthetic 360° dataset and trained the model with AdamW~\cite{loshchilov2017decoupled} and a learning rate of 1e-4 for the LoRA layers, using one NVIDIA A100 GPU and a batch size of \(4\). We trained for 100K iterations, as further fine-tuning caused noticeable identity loss, making it harder for SDS to handle these inconsistencies.

\subsection{FLAME-based Point Cloud Initialization}
\label{sec:implement_init}

We initialize the splats based on the FLAME mesh, which consists of \( N_{\text{original}} = 5023 \) vertices. However, given the low vertex count, we first perform dense sampling of the mesh. Maintaining consistency in the upsampling process is essential to ensure that when expression blendshapes are applied to the facial region, the resulting deformations are consistently upsampled and accurately incorporated into the upsampled facial mesh. To achieve this, we apply the \texttt{subdivide()} method from the \texttt{trimesh} \cite{trimesh} library, which implements the Midpoint Subdivision algorithm. This process upscales the original mesh to \( N_{\text{upsampled}} = 79936 \) vertices, with the majority concentrated in the facial area of interest (\( N_{\text{face}} = 70033 \) vertices) and the remaining \( N_{\text{head}} = 9903  \) vertices allocated to the rest of the head. Since we are only concerned with maintaining consistent upsampling within the facial region, we separate the mesh into facial and head components. The head component is then independently upsampled to \( N_{\text{head}} = 73050 \) vertices. Finally, the facial and head meshes are reconnected, resulting in a unified point cloud that serves as the initialization for the optimized splat \( G_{\text{init}}(\mathbf{x}) \).

\subsection{SDS-ISM Parameters}
\label{sec:implement_init}

Our distillation framework is based on ISM \cite{ISM}. The settings detailed below are presented in full correspondence with their method, following the same format.

\subsubsection{Guidance Parameters}

As discussed in the main paper, our strong prior and carefully designed task-specific SDS process, along with settings refined through experimentation, eliminate the need for the high guidance scale typically associated with SDS approaches, effectively avoiding color issues. Specifically, we employ a scale of \(1\) and the Perp-Neg algorithm \cite{armandpour2023re}, and, following the notation of \cite{ISM}, we use \(\delta_T = 40\) paired with \(\delta_S = 20\), utilizing \(20\) inversion steps. This 
results in 3D avatars that exhibit high detail and natural color.

\subsubsection{Camera Parameters}

We utilize a random camera sampling strategy with progressively relaxed view ranges during training. The initial camera configurations are:

\begin{itemize}
    \item Radius range: \([5.2, 5.5]\).
    \item Maximum radius range: \([4.2, 5.2]\).
    \item Field of view (FoV) range: \([0.53, 0.53]\).
    \item Maximum FoV Range: \([0.3, 0.7]\).
    \item Elevation angle range (\(\theta\)): \([40^\circ, 100^\circ]\).
    \item Azimuth angle range (\(\phi\)): \([-180^\circ, 180^\circ]\).
\end{itemize}

Starting from iteration \(2000\), we progressively relax the camera view ranges every \(2000\) iterations by scaling the parameters:

\begin{itemize}
    \item FoV factor: \([0.8, 1.1]\).
    \item Radius factor: \(0.95\).
\end{itemize}

\subsubsection{Optimization Parameters}

We train our avatars with a rendering resolution of \(512 \times 512\) pixels for \(6000\) iterations on a single NVIDIA RTX 4090 GPU (24GB) using a batch size equal to \(4\). Optimizing an avatar for an input subject takes approximately 80 minutes, and the final avatar typically consists of nearly 110K Gaussians.

The optimization is performed using the Adam optimizer with \(\beta_1 = 0.9\), \(\beta_2 = 0.999\), and \(\epsilon = 1\mathrm{e}{-15}\). The learning rates for different parameters are scheduled to decay exponentially from their initial values to final values over the course of training, using a delay multiplier of \(0.01\):

\begin{itemize}
    \item Position (\(\bm{\mu}\)): \(\text{lr}_{\text{init}} = 1.6\mathrm{e}{-4}\), \(\text{lr}_{\text{final}} = 1.6\mathrm{e}{-6}\).
    \item Color (\(\mathbf{f}\)): \(\text{lr}_{\text{init}} = 5\mathrm{e}{-3}\), \(\text{lr}_{\text{final}} = 3\mathrm{e}{-3}\).
    \item Opacity (\(\bm{\alpha}\)): \(\text{lr} = 5\mathrm{e}{-2}\).
    \item Scaling (\(\mathbf{s}\)): \(\text{lr}_{\text{init}} = 5\mathrm{e}{-3}\), \(\text{lr}_{\text{final}} = 1\mathrm{e}{-3}\).
    \item Rotation (\(\mathbf{r}\)): \(\text{lr}_{\text{init}} = 1\mathrm{e}{-3}\), \(\text{lr}_{\text{final}} = 2\mathrm{e}{-4}\).
\end{itemize}

\subsection{Splat Modification Strategy}

To refine the splats in the non-facial areas, we initiate densification and pruning at iteration \(1000\), performing them every \(500\) iterations until \(5000\). During this period, opacity resets are also applied every \(1000\) iterations. In the final \(1000\) iterations, we further refine the splats by pruning disconnected splats every \(100\) iterations to remove isolated noise and applying targeted pruning based on opacity and size every \(200\) iterations.

\subsection{Camera Sampling Strategy}

Given the approximate symmetry of human heads, we observed that sampling an equal number of front and back views during training was more beneficial than randomly sampling any azimuth angle. To achieve this, we enforced the sampling of four azimuth angles for each training step:

\begin{itemize}
    \item Two angles from the frontal range \([-90^\circ, 90^\circ]\):
    one from \([-90^\circ, 0^\circ)\) and one from \([0^\circ, 90^\circ)\).
    \item Two angles from the back range \([-180^\circ, -90^\circ) \cup (90^\circ, 180^\circ]\):
    one from \([-180^\circ, -90^\circ)\) and one from \((90^\circ, 180^\circ]\).
\end{itemize}

This strategy ensured a balanced and diverse set of views, encompassing frontal, back, and side perspectives. 

\subsection{Template Regularization}
As discussed in the main paper, we employ strong template proximity regularizers, specifically the \(L_2\) distance regularizer and the Laplacian difference regularizer. Through experimentation, we found that using high weights for these regularizers leads to very strong template correspondence. Therefore, we selected a value of \(1\mathrm{e}{+8}\) for both.

\end{document}